\pdfoutput=1

\documentclass[11pt]{article}

\usepackage[final]{acl}

\usepackage{times}
\usepackage{latexsym}
\usepackage{amsmath}

\usepackage[T1]{fontenc}

\usepackage[utf8]{inputenc}

\usepackage{microtype}

\usepackage{inconsolata}

\usepackage{graphicx}
\usepackage{amsmath}

\usepackage[inline]{enumitem}

\usepackage[linesnumbered,ruled,vlined]{algorithm2e}
\usepackage{hyperref}
\usepackage{tablefootnote}
\usepackage{colortbl}
\usepackage{graphicx}
\usepackage{multirow}

\usepackage{booktabs}
\usepackage{stfloats} 

\usepackage[normalem]{ulem}

\title{Cross-table Synthetic Tabular Data Detection}

\author{
 \textbf{G. Charbel N. Kindji\textsuperscript{1,2}},
 \textbf{Lina M. Rojas Barahona\textsuperscript{1}},
 \textbf{Elisa Fromont\textsuperscript{2}},
 \textbf{Tanguy Urvoy\textsuperscript{1}},
\\
\\
 \textsuperscript{1}Orange Labs Lannion \\ 
    \small{\texttt{first.last@orange.com}}, \\
 \textsuperscript{2}Université de Rennes, CNRS, Inria, IRISA UMR 6074 \\       \small{\texttt{first.last@irisa.fr}}
\\
 \small{
    \textbf{Correspondence:} \href{mailto:charbel.kindji.orange@gmail.com}{charbel.kindji.orange@gmail.com}
 }
}

\begin{document}
\maketitle
\begin{abstract}

Detecting synthetic tabular data is essential to prevent the distribution of false or manipulated datasets that could compromise data-driven decision-making. This study explores whether synthetic tabular data can be reliably identified "in the wild"—meaning across different generators, domains, and table formats. This challenge is unique to tabular data, where structures (such as number of columns, data types, and formats) can vary widely from one table to another. We propose three cross-table baseline detectors and four distinct evaluation protocols, each corresponding to a different level of "wildness". Our very preliminary results confirm that cross-table adaptation is a challenging task.
\end{abstract}

\section{Introduction and Related Works}

Most studies on synthetic data detection focus on image~\cite{chai2020makes, corviDetectImg2023,marraGanFingerprints2019,BammeySynthbuster}, text~\cite{lavergne2011filtering, lahby2022online, hu2023radar,wang-etal-2024-ideate,MitchellDetectGPT}, audio~\cite{lopez2016revisiting}, video (face-swap)~\cite{pu2021deepfake}, or their combination~\cite{singhal2020spotfake+}.

Nevertheless, a growing number of generative models for tabular data generation has emerged recently; some are general-purpose~\cite{zhang2023mixed, kotelnikov2023tabddpm}, while others are tailored to specific domains like finance~\cite{findiff2023} or healthcare~\cite{Hyun2020ASD}.
With these advances it will be easier to generate realistically manipulated datasets to fake scientific results or to hide fraud and accouting loopholes.
It is therefore essential to focus research efforts on the detection of synthetic tabular data, and to develop detection techniques that are on par with the impressive generative models' capabilities.

Detecting syntetic content issued from a known generative model on a restricted domain is a fairly tractable task. The performance of such a predictor is indeed  commonly used for adversarial training~\cite{goodfellow2020generative} and as a metric to assess generation performance~\cite{lopez2016revisiting, c2st22}.

However, the challenge intensifies when attempting to detect synthetic data "in the wild" \cite{stadelmann2018deepwild}, namely, 
when the deployed system has to face modalities and content generators it has never seen during its training phase. Is is known that, even for homogeneous formats like image or text, synthetic content detection systems are not robust to  such \emph{cross-generator} and \emph{cross-domain} distribution shifts \cite{kuznetsov2024robust}.

When dealing with tabular data, we have to face a stronger form of domain-shift that we call \emph{cross-table} shift. Indeed, for a synthetic table detection system to be useful, is has to cope with different table formats with varying numbers of columns, varying types and varying distributions shapes.
Although, the litterature on domain adaptation across the same table structure is vast \cite[see][for a survey]{gardner2024benchmarking}, only a few recent articles propose classifiers that are able to generalize across different tables \cite{wang2022transtab,spinaci2024portal}. To the best of our knowledge, no study on cross-table synthetic data detection has been published yet.

We present a preliminary work with three baselines for synthetic tabular data detection "in the wild." We focus on cross-table robustness among different  real-world evaluation scenarios representing various degrees of "wildness", for instance:
\begin{enumerate*}[label=(\roman*)]
    \item \textit{No shift:} the model is trained and tested on samples from the same pool of datasets and generators;
    \item \textit{Cross-generator shift:} the model is tested on the same datasets but the test synthetic data is produced by unknown generators;
    \item \textit{Cross-table shift:} the model is tested on holdout datasets and table structures but with synthetic data produced by known generators;
    \item \textit{Full shift:} the model is tested on generators and datasets it has never seen before.
\end{enumerate*}

We address here the cross-table adaptation by considering two \emph{text-based} baselines where the table rows are first linearized as strings, and a \emph{table-based} transformer with a simple column-wise table-agnostic encoding.

\section{Real and Synthetic Data}
\label{s:data}
\paragraph{Real Data:} We use 14 common public tabular datasets from the UCI\footnote{\url{https://archive.ics.uci.edu/}} with different sizes, dimensions and domains. These datasets are described in Table~\ref{tab:datasets}.

\setlength{\abovecaptionskip}{5pt} 
\setlength{\belowcaptionskip}{-15pt} 
\begin{table}[hptb]
\small
\centering
    \begin{tabular}{cccc}\hline
        Name & Size & \#Num &  \#Cat\\\hline
        Abalone\tablefootnote{\label{fn:dataset_link_openml}\url{https://www.openml.org}} & 4177 & 7 & 2\\
        Adult\footref{fn:dataset_link_openml} & 48842 & 6 & 9 \\
        Bank Marketing\footref{fn:dataset_link_openml} & 45211 & 7 & 10 \\
        Black Friday\footref{fn:dataset_link_openml} & 166821 & 6 & 4 \\
        Bike Sharing\footref{fn:dataset_link_openml} & 17379 & 9 & 4 \\
        Cardio\tablefootnote{\label{fn:dataset_link_kaggle}\url{https://www.kaggle.com/datasets}} & 70000 & 11 & 1 \\
        Churn Modelling\footref{fn:dataset_link_kaggle} & 4999 & 8 & 4 \\
        Diamonds\footref{fn:dataset_link_openml} & 26970 & 7 & 3 \\
        HELOC\footref{fn:dataset_link_kaggle} & 5229 & 23 & 1 \\
        Higgs\footref{fn:dataset_link_openml} & 98050 & 28 & 1 \\
        House 16H\footref{fn:dataset_link_openml} & 22784 & 17 & 0 \\
        Insurance\footref{fn:dataset_link_kaggle} & 1338 & 4 & 3 \\
        King\footref{fn:dataset_link_kaggle} & 21613 & 19 & 1 \\
        MiniBooNE\footref{fn:dataset_link_openml} & 130064 & 50 & 1 \\
    \end{tabular}
    \caption{Description of the datasets. "\#Num" refers to the number of numerical attributes and "\#Cat" the number of categorical ones.}
    \label{tab:datasets}
\end{table}

\paragraph{Synthetic Data: } Our \emph{data generators} are heavily tuned versions of  TabDDPM~\cite{kotelnikov2023tabddpm}, TabSyn~\cite{zhang2023mixed}, TVAE, and CTGAN~\cite{CTGAN} provided by~\cite{kindji2024hoodtabulardatageneration}. We trained the models on the entire real datasets before sampling new synthetic rows. Each model is used to create a synthetic version of each dataset. 

\section{Detection Models}
\label{sec:detection_models}

In order to be useful "in the wild", a detection model must be "table-agnostic", which means that  it must accept inputs form different table formats.
We trained three baselines for synthetic content detection from scratch: a {logistic regression} and two transformer-based classifiers.
For the {logistic regression} and the first transformer the table is first linearized into text (Section \ref{sec:text-based}). For the second transformer-based classifier we use a rough columns level encoding of tables (Section \ref{sec:table-based}). 

The transformer-based classifiers have three main components: \begin{enumerate*}[label=(\roman*)] \item a feature embedding block, \item a transformer encoder block, and \item a classification head. \end{enumerate*}
As in BERT, the classifier relies on a \textit{CLS} embedding that is added to the input and retrieved in the output of the transformer blocks. The \textit{CLS} representation is fed to the classification head to predict the binary target class (real or synthetic data). The models (both \textit{text-based} and \textit{table-based}) are trained using a binary cross entropy loss.

\subsection{Text-Based Encodings}
\label{sec:text-based}

A natural solution to build a table-agnostic model is to consider the tables as raw text.
This approach is used in pretrained models such as TaBERT~\cite{yin20acl}, TAPAS~\cite{herzig-etal-2020-tapas}, or TAPEX~\cite{liu2022tapex}.
These models are are designed to encode small tables like the ones found on Wikipedia. They are derived from BERT and rely on a text encoding of the whole table.

In order to work with larger tables we opted, as in \cite{borisovlanguage23}, to work at the row level.
We converted each table row into a shuffled sequence of \texttt{<column>:<value>} patterns.

For instance the first row of Table~\ref{tab:datasets} can be encoded as the string
\texttt{"Name:Abalone,Size:4177,\#Num:7,\#Cat:2"} or any of its column permutations. This random columns' permuation is intended to increase generalization across different tables.
Then two options are considered:
\begin{enumerate*}[label=(\roman*)]
\item For the logistic regression, the string is simply split into a bag of character-level trigrams like \texttt{"Nam", "e:A", ":41"} or ,\texttt{"t:2"};
\item For the text-transformer baseline the string is tokenized into a sequence of characters that are mapped, as usual for transformers, into a sequence of embedding vectors that are combined with a positional embedding. 
\end{enumerate*}

\subsection{Table-Based Encodings}
\label{sec:table-based}

All datasets are encoded following the same procedure: 
numerical features are normalized through \textit{QuantileTransformer}, and categorical features are encoded with the \textit{OrdinalEncoder}, both from scikit-learn\footnote{\href{https://scikit-learn.org/stable/}{https://scikit-learn.org/stable/}}. Importantly, each dataset is processed separately. This means that the methods used to encode numerical and categorical features are applied to each dataset individually, rather than collectively. 
The feature embedding module employs a shared feed-forward layer for numerical features and a shared embedding layer for categorical features. This baseline is of course simplistic, more sophisticated strategies are proposed in \cite{wang2022transtab} and \cite{spinaci2024portal}.

\section{Experimental Setup}
\label{sec:exps}

All dataset rows are mixed together in a list with two additional labels: the \textit{dataset name} and the \textit{origin} that can be \textit{"real"} or the name of its generator if the row is synthetic.
We use these two additonal labels to design cross-validation splits with increasingly challenging constraints:
\vspace{-1em}
\begin{equation*}
    \begin{array}{l}
    \text{Generator:}\left\{
    \begin{array}{l} 
      \text{Single}\\
      \text{Multiple}, \text{Cross-generator}\\
    \end{array}
      \right.\\
  \text{Table:} \left\{
        \begin{array}{l} 
          \text{Single}\\
        \text{Multiple}, \text{Cross-table}
    \end{array}\right.\\
          
    \end{array}
\end{equation*}
\vspace{-1em}

For instance, the \textit{Classifier Two-Samples Test} (C2ST) metric as described in \cite{lopez2016revisiting, c2st22} correspond to the simplest \textit{Single Generator vs Real, Single Table} setting. It does not require a "table-agnostic" model.
The \textit{cross-generator shift} constraint guarantees that a generator used for trainning cannot be used in test. The cross-table constraint guarantees that a table used for trainning cannot be used in test. These single-criterion shift settings can be coded using Scikit-Learn \textit{GroupKFold}. However, as shown in Table~\ref{tab:drift_combination_example}, cross-validating a \textit{Full shift} with both cross-table and cross-generator robustness is a bit trickier.

\begin{table}[htbp]
\small
\centering
\begin{tabular}{ll|ccl|}
\cline{3-5}
                                                            &   & \multicolumn{3}{c|}{\textbf{Tables}}                                                                                   \\ \cline{3-5} 
                                                            &   & \multicolumn{1}{l|}{A}                        & \multicolumn{1}{l|}{B}                        & C                        \\ \hline
\multicolumn{2}{|c|}{\textit{Real Data}}                        & \multicolumn{1}{l|}{\cellcolor[HTML]{D0E7FF}} & \multicolumn{1}{l|}{\cellcolor[HTML]{DFFFD6}} & \cellcolor[HTML]{D0E7FF} \\ \hline
\multicolumn{1}{|l|}{}                                      & X & \multicolumn{1}{c|}{\cellcolor[HTML]{D0E7FF}} & \multicolumn{1}{c|}{\cellcolor[HTML]{D3D3D3}} & \cellcolor[HTML]{D0E7FF} \\ \cline{2-5} 
\multicolumn{1}{|l|}{}                                      & Y & \multicolumn{1}{c|}{\cellcolor[HTML]{D3D3D3}} & \multicolumn{1}{c|}{\cellcolor[HTML]{DFFFD6}} & \cellcolor[HTML]{D3D3D3} \\ \cline{2-5} 
\multicolumn{1}{|l|}{\multirow{-3}{*}{\textbf{Generators}}} & Z & \multicolumn{1}{c|}{\cellcolor[HTML]{D0E7FF}} & \multicolumn{1}{c|}{\cellcolor[HTML]{D3D3D3}} & \cellcolor[HTML]{D0E7FF} \\ \hline
\end{tabular}
\caption{Example of a full shift split. The blue cells indicate the training elements, while the green cells represent the test sets. The gray cells indicate examples that must be dropped because they would violate one of the Tables or Generator separation constraints.
}
\label{tab:drift_combination_example}
\end{table}

\subsection{Detection Without Distribution Shift}
\label{sec:no_drift}

We first train models to detect synthetic data generated only by \textit{TVAE}~\cite{CTGAN}. Despite our interest in "model agnostic" detection, this procedure provides an upper-bound reference to compare with.
This setup is referred as \textit{TVAE vs Real, All-Tables, No Shift}.
We then add an additional setup where synthetic datasets from all models are mixed to be detected against the real datasets. We refer to this setup as \textit{All Models vs Real, All-Tables, No Shift}. 

\subsection{Detection Under Distribution Shifts}
\label{sec:data_drift_setup}

\begin{table}[htpb]
\small
\centering

\begin{tabular}{ll|ccl|}
\cline{3-5}
                                                            &   & \multicolumn{3}{c|}{\textbf{Tables}}                                                                                   \\ \cline{3-5} 
                                                            &   & \multicolumn{1}{l|}{A}                        & \multicolumn{1}{l|}{B}                        & C                        \\ \hline
\multicolumn{2}{|c|}{\textit{Real Data}}                        & \multicolumn{1}{l|}{\cellcolor[HTML]{D0E7FF}} & \multicolumn{1}{l|}{\cellcolor[HTML]{D0E7FF}} & \cellcolor[HTML]{DFFFD6} \\ \hline
\multicolumn{1}{|l|}{}                                      & X & \multicolumn{1}{c|}{\cellcolor[HTML]{D0E7FF}} & \multicolumn{1}{c|}{\cellcolor[HTML]{D0E7FF}} & \cellcolor[HTML]{DFFFD6} \\ \cline{2-5} 
\multicolumn{1}{|l|}{}                                      & Y & \multicolumn{1}{c|}{\cellcolor[HTML]{D0E7FF}} & \multicolumn{1}{c|}{\cellcolor[HTML]{D0E7FF}} & \cellcolor[HTML]{DFFFD6} \\ \cline{2-5} 
\multicolumn{1}{|l|}{\multirow{-3}{*}{\textbf{Generators}}} & Z & \multicolumn{1}{c|}{\cellcolor[HTML]{D0E7FF}} & \multicolumn{1}{c|}{\cellcolor[HTML]{D0E7FF}} & \cellcolor[HTML]{DFFFD6} \\ \hline
\end{tabular}
\caption{Example of a \textit{cross-table shift} split. The blue cells indicate the training elements, while the green cells represent the test set.}
\label{tab:dataset_drift_example}
\end{table}

We have tested our baselines only under the \textit{cross-table} shift constraint, which proves to be already quite challenging.
As illustrated in Table~\ref{tab:dataset_drift_example}, in this scenario the detection model is first trained on real and synthetic datasets produced by some generators and then deployed on unseen datasets. 

\section{Results}
\label{sec:results}

In this section, we present our baselines' results on different setups, without and with \textit{cross-table shift}. These results are summarized in Table~\ref{tab:results} with the standard \textit{ROC-AUC} and \textit{Accuracy} metrics.

\renewcommand{\arraystretch}{1.05} 
\setlength{\tabcolsep}{3pt} 
\setlength{\abovecaptionskip}{5pt} 
\setlength{\belowcaptionskip}{-15pt} 

\begin{table}[ht]
    
    \centering
    \small
    \begin{tabular}{@{}clcc@{}}
        \toprule
        \multirow{2}{*}{Setup} & \multirow{2}{*}{Model} & \multicolumn{2}{c}{Metrics} \\ 
        \cline{3-4}
                               &                         & AUC & Accuracy  \\ 

        \midrule
        \multirow{2}{*}{\shortstack[c]{TVAE vs Real, \\ All Tables, \\ No shift}}
        & 3grm-LReg.  & $0.71$ & $0.65$ \\ 
        \cline{2-4}
        & Text-Transf. & $0.76$ & $0.68$ \\ 
        \cline{2-4}
        & Table-Transf. & {$\boldsymbol{0.91}$} & {$\boldsymbol{0.82}$} \\
        
        \midrule
        \midrule
       
        \multirow{2}{*}{\shortstack[c]{All Models vs Real, \\ All Tables, \\ No shift}} 
        & 3grm-LReg. & $0.67$ & $0.62$ \\ 
        \cline{2-4}
        & Text-Transf. & {$\boldsymbol{0.78}$} & {$\boldsymbol{0.72}$} \\ 
        \cline{2-4}
        & Table-Transf.  & $0.77$ & $0.69$ \\
        \midrule
        \midrule
        \multirow{2}{*}{\shortstack[c]{All Models vs Real, \\ All Tables, \\ Cross-table shift}} 

        & 3grm-LReg. & {$\boldsymbol{0.58}$} & {$\boldsymbol{0.55}$} \\
        \cline{2-4}
        & Text-Transf. & 0.56 & 0.52 \\ 
        \cline{2-4}
        & Table-Transf. & 0.51 & 0.50 \\ 
        \bottomrule
    \end{tabular}
    \caption{Evaluation of synthetic tabular data detection on various setups. "3grm-LReg." stands for "Trigrams Logistic Regression" and "Transf." stands for "Transformer"}
    \label{tab:results}
\end{table}

\subsection{Without Distribution Shift}
\label{sec:results_no_drift}

The transformer-based models (both \textit{text-based} and \textit{table-based}) demonstrate good performance across various metrics, under both setups \textit{TVAE vs Real} and \textit{All models vs Real}. 
We notice an \textit{AUC} over $0.76$ for all setups suggesting a good generalization capabilities of these table-agnostic models. %
Despite its rather naive design, the \textit{AUC} for detecting TVAE-generated rows of our table-agnositc tranformer baseline reachs $0.91$. It is worth comparing this result with the ones obtained in single dataset settings: in~\cite{kindji2024hoodtabulardatageneration} the \textit{XGBoost} \textit{TVAE~vs~Real} median \textit{AUC} for detecting TVAE is $0.81$.

The task difficulty increases under the \textit{All models vs Real} setup, but the overall performance remains stable for all models.
The \textit{table-based} transformer outperforms the \textit{text-based} version in \textit{TVAE vs Real}, however, it underperforms in \textit{All Models vs Real}. Note that the only difference between the two approaches lies in the preprocessing and the way the feature embedding module  works (as detailed in Sections \ref{sec:table-based} and \ref{sec:text-based}). This suggests that the textual representation offers a more general view across all models and datasets. 
As a side result, we notice that there is still significant room for improvement in achieving realism in tabular data generation. The synthetic tabular data generators seems to exhibit patterns that a naive table-agnostic classifier is able to detect.

\subsection{Cross-table Shift}
\label{sec:results_drift}

The \textit{cross-table shift} results (Table~\ref{tab:results}) show that this setup is particularly challenging, as all models struggle to achieve good performance. The \textit{table-based} approach drops significantly  its performance (\textit{AUC}$=0.51$). The model fails to identify meaningful patterns and cannot generalize to unseen datasets, essentially making random guesses on the test set.

An interesting observation is that the \textit{text-based} transformer appears to provide more generalizable patterns than the \textit{table-based} one. This aligns with the results from the \textit{All Models vs Real} setup, in which it also performed better. As there are more datasets and models to generalize across, this approach benefits from that diversity. However, the \textit{AUC} score is relatively low at $0.56$.
The training curves presented in Appendix~\ref{sec:appendix_additional_results} confirms that, with a \textit{cross-table shift} between all training, validation, and test sets; the text-based transformer (on the left-hand side) is more robust than the \textit{table-based} transformer (on the right-hand side).
The \textit{dataset-agnostic} encoding we used in the \textit{table-based} method reveals its limitations when evaluated on unseen datasets. Being tied to datasets particularities, the encoding do not generalize well to datasets with different characteristics (e.g. the number of features, range of numerical features, categories in categorical features, and sample size). In contrast, the textual representation captures patterns that can be generalized.

As expected, due to its extreme simplicity, the logistic regression model outperformed the transformers for the \textit{cross-table shift} setup with an AUC at $0.58$ (versus $0.56$ for the \textit{text-based} transformer). However,  an AUC of $0.58$ is not a very impressive result and, contrary to transformers~\cite{zhou2024what, li2023systematic, yadlowsky2024can}, its potential for improvement is weak.

These preliminary results suggests further investigations on transformer-based models with both text-based and table-based encodings.
The potential for transfer learning from pre-trained models can also enhance performance, making transformer-based approaches a valuable asset in the \textit{cross-table shift} setup.

\section{Conclusion}
\label{sec:concl}
We study synthetic tabular data detection "in the wild". We utilized $14$ datasets and $4$ state-of-the-art, highly-tuned tabular data generation models. We evaluated various models using different tabular data representations as inputs and demonstrated that it is possible to detect synthetic data with promising performance. We also introduced various levels of "wildness" that correspond to different degrees of data distribution shift and we focused on \textit{cross-table shift}. Our preliminary results are encouraging but show that cross-table adaptation is still a challenging problem. In the future, we will consolidate these results and explore more sophisticated encodings and adaptation strategies such as including table metadata---like column names---in the input. We also plan to explore the adaptation of pretrained encoders like TaBERT to see if they reach the performance of our baselines on fake content detection. 

\section{Limitations}
As the results showed, the \textit{table-based} transformer, along with its preprocessing and feature embedding scheme, provides valuable insights when there is no distribution shift. However, it struggles to generalize when a \textit{cross-table shift} is introduced. We believe this encoding scheme has the merit of its simplicity, but it needs to be enhanced for distribution shift scenarios by incorporating general dataset information, such as column names and category embeddings as it is done in \cite{spinaci2024portal}. These improvements should help differentiate between synthetic and real data if synthetic data fails to accurately replicate these characteristics.
On the other hand, the textual encoding offers the advantage of being simpler and more general, but it leads to longer row-encoding sequences and it lacks of a tabular-specific inductive bias.

We implemented straightforward baselines utilizing both common NLP techniques and transformer architecture. 
For now, we did not conduct ablation studies to examine the impact of input column permutation and positional encoding. We also did not consider other table format specificities such as table size, number of columns, and data types.

The few experiments we did to adapt TaBERT on larger tables were not conclusive. We suspect, that BERT-like tokenization and small tables pre-training is not adapted to our problem, but it requires further investigations that we keep for future work. 

\bibliography{biblioCharbel}

\appendix

\section{Additional Distribution Shifts}
\label{sec:appendix_drifts}

We explore several challenging distribution shift setups for evaluating synthetic tabular detection data "in the wild". We evaluated our baselines on the \textit{cross-table shift} and provide additional information about the remaining distribution shifts setups. 
\subsection{Cross-generator Shift}

As illustrated in Table\ref{tab:generator_drift_example}, for generator shift, the model is trained to distinguish between real and synthetic data from some generators and some datasets. The model is then tested with synthetic data produced by generators it has never seen before.

\begin{table}[htpb]
\small
\centering
\begin{tabular}{ll|ccl|}
\cline{3-5}
                                                            &   & \multicolumn{3}{c|}{\textbf{Tables}}                                                                                   \\ \cline{3-5} 
                                                            &   & \multicolumn{1}{l|}{A}                        & \multicolumn{1}{l|}{B}                        & C                        \\ \hline
\multicolumn{2}{|c|}{\textit{Real Data}}                        & \multicolumn{1}{l|}{\cellcolor[HTML]{D0E7FF}} & \multicolumn{1}{l|}{\cellcolor[HTML]{D0E7FF}} & \cellcolor[HTML]{DFFFD6} \\ \hline
\multicolumn{1}{|l|}{}                                      & X & \multicolumn{1}{c|}{\cellcolor[HTML]{D0E7FF}} & \multicolumn{1}{c|}{\cellcolor[HTML]{D0E7FF}} & \cellcolor[HTML]{D0E7FF} \\ \cline{2-5} 
\multicolumn{1}{|l|}{}                                      & Y & \multicolumn{1}{c|}{\cellcolor[HTML]{D0E7FF}} & \multicolumn{1}{c|}{\cellcolor[HTML]{D0E7FF}} & \cellcolor[HTML]{D0E7FF} \\ \cline{2-5} 
\multicolumn{1}{|l|}{\multirow{-3}{*}{\textbf{Generators}}} & Z & \multicolumn{1}{c|}{\cellcolor[HTML]{DFFFD6}} & \multicolumn{1}{c|}{\cellcolor[HTML]{DFFFD6}} & \cellcolor[HTML]{DFFFD6} \\ \hline
\end{tabular}
\caption{Example of \textit{cross-generator shift} split. The blue cells indicate the training elements, while the green cells represent the test set. Here, all rows associated with generators X and Y were selected for the train set. Note that there are some \textit{real} datasets in the training set as well.}
\label{tab:generator_drift_example}
\end{table}

\subsection{Full Shift}

Another critical scenario arises when the model is trained on a specific set of generators and datasets, but encounters unseen generators and datasets during deployment. Here there is a \textit{cross-table shift} and a \textit{cross-generator shift}. In this scenario, the model could struggle to generalize learned patterns to totally unseen data. The schematic representation is provided in Table \ref{tab:drift_combination_example}. Due to the constraints on the datasets and generators in this setup, certain data cannot be included in either the training set or the test set.

\section{Additional Results}
\label{sec:appendix_additional_results}

\begin{figure*}[t]
  \includegraphics[width=0.48\linewidth]{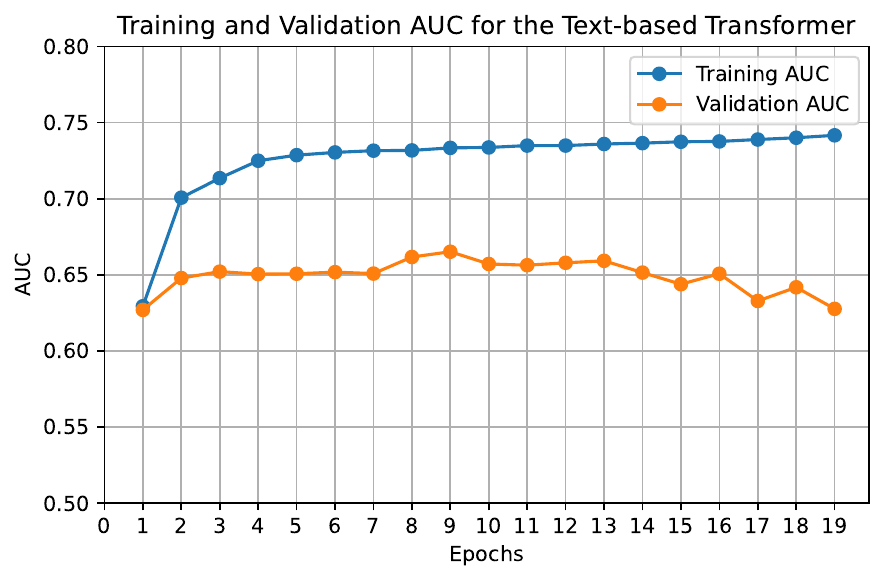} \hfill
  \includegraphics[width=0.48\linewidth]{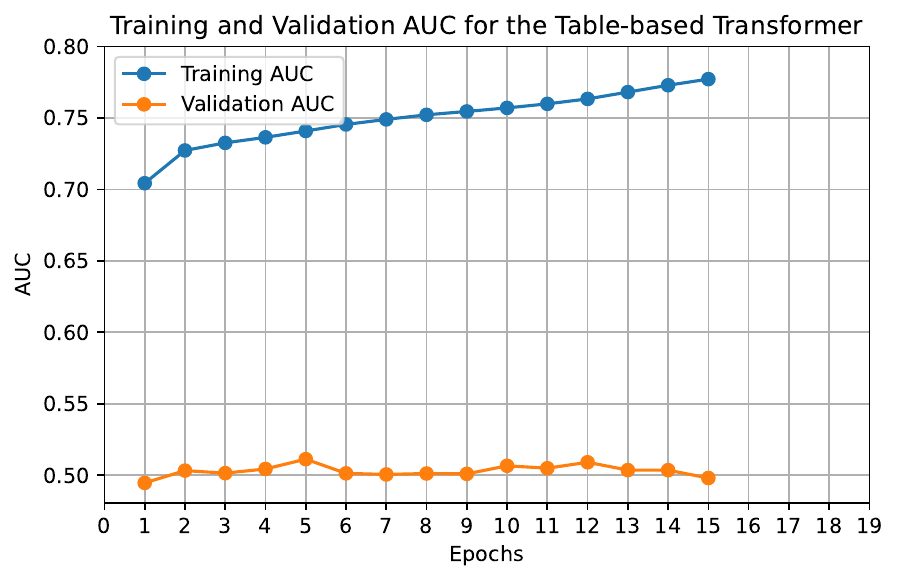}
  \caption {Training and validation \textit{AUC} performance of models trained under \textit{cross-table shift} setup. Left: \textit{text-based} model and right: \textit{table-based} approach.}
  \label{fig:train_val_auc}
\end{figure*}

We provide the training and validation curves for the \textit{AUC} metric for the \textit{cross-table shift} setup in Figure~\ref{fig:train_val_auc}. 

\end{document}